# Estimation of English and non-English Language Use on the WWW

**Gregory Grefenstette & Julien Nioche**
Xerox Research Centre Europe
6 chemin de Maupertuis,
38240 Meylan, France
{grefen, jnioche}@xrce.xerox.com

**Abstract**

The World Wide Web has grown so big, in such an anarchic fashion, that it is difficult to describe. One of the evident intrinsic characteristics of the World Wide Web is its multilinguality. Here, we present a technique for estimating the size of a language-specific corpus given the frequency of commonly occurring words in the corpus. We apply this technique to estimating the number of words available through Web browsers for given languages. Comparing data from 1996 to data from 1999 and 2000, we calculate the growth of a number of European languages on the Web. As expected, non-English languages are growing at a faster pace than English, though the position of English is still dominant.

**Introduction**

In their attempt to understand the anarchically expanding World Wide Web, researchers have been trying to estimate a variety of Web characteristics (Bray 96; Woodruff 96; Pitkow 98; Trau 99). Some characteristics are easy to obtain. For example, by polling the list of registered network addresses, 56 million hosts were counted in July 1999, and 72 million were counted in January 2000. A plot[1] of the growth of the Web in terms of computer hosts can therefore be generated from such historical data. Other characteristics are harder to discern, and can only be estimated by sampling and extrapolation. Lawrence and Gilles (1999) estimated in February, 1999, that there were 800 million indexable web pages available, by comparing the overlap[2] between page lists returned by different web browsers over the same set of queries. By sampling pages, and estimating an average page length of 7 to 8 kilobytes of non-HTML text, they concluded that there might exist 6 terabytes of indexable text on the WWW. Another interesting characteristic of the Web was found by Lada Adamic (1999). Working from a 1997 snapshot of the Web, she estimated that the Web possesses the Small World property: from any web-site connected to another web-site, the average number of links to get from one to the other is 4.2 links.

Beyond these structural characteristics of the Web pages and servers, there is a great interest in characterizing the textual content of the WWW. One of the most easily remarkable aspects of the WWW from a content perspective is its multilinguality (Nunberg, 1996). A number of studies of language use, often involving some type of automatic language detection (Grefenstette, 1995), have been performed. David Crystal (1997) retrieved pages containing international words (such as *Internet*) and calculated that about 80% of those pages were English. In June 1997, the Babel Team retrieved 3239 home pages[3] by randomly generating IP addresses and then trying to access them as if they were URLs. Each page's language was automatically identified and then verified manually. They found that English was the principal language on 82.3% of these pages, followed by German (4.0%), Japanese (3.1%), French (1.8%) and Spanish (1.1%). This test supposed that home pages are representative of the Web, and that IP addresses are randomly distributed over the world.

---

[1] http://www.isc.org/ISC/news/pr2-10-2000.html
[2] http://www.neci.nec.com/~lawrence/websize.html
[3] See http://babel.alis.com:8080/palmares.html

Geoffrey Nunberg and Hinrich Schütze (1998) also estimated the amount of English in top-level Internet hosts for major linguistic communities. 2.5 million Web pages were drawn from a 1997 Internet Archive Web crawl and sorted by host-machine domain (e.g., *.bg* is used for Bulgarian hosts, *.cn* is used for Chinese hosts, etc.) Each page was passed through an automatic language identifier. This report revealed a high use of English on pages in developing countries: e.g., 84% of the top-level pages in hosts in Romania were written in English, 76% of top-level pages in Latvia. In small, developed language communities such as Sweden 40% of top-level pages were still written in English, as were 46% of the top-level pages in the Denmark domain (*.dk*). Jack Xu (2000) reports on experiments applying a language identifier to web pages indexed by the Excite portal[4] that finds that 71% of the pages (453 million out of 634 million web pages) are written in English. About 29% is written in languages other than English. Breaking down the pages by language gives English (71%), Japanese (6.8%), German (5.1%), French (1.8%), Chinese (1.5%), Spanish (1.1%), Italian (0.9%), and Swedish (0.7%).

Here in this paper we report on similar work that estimates the presence of various languages on the Web. Rather than counting or estimating a number of Web pages, however, we present a technique for estimating the raw number of words that are accessible for a given language through a web portal. The next section presents our method for estimating the size in words of an unknown corpus based on the known frequency of a limited number of words. The following section shows how this estimation technique can use frequency information provided by Web portals to estimate language size, in words, on the Web.

## Method and Local Evaluation

In order to understand the method we present, consider the following question. Suppose that you only know the frequency of one word in a text corpus, is it possible to guess the size of the entire corpus?

**Anecdotal Example**
The most commonly used word in written English is the word *the*. In the 90-million word corpus of written English from the British National Corpus[5], the word *the* appears 5,776,487 times. This is roughly 7 times for every 100 words. Now, suppose that we know how many times the word *the* appears in a text, without knowing how long the text is. For a concrete example, in the American Declaration of Independence, *the* occurs 84 times. Knowing that *the* appears seven times in every one hundred words, we predict that the Declaration is about 1200 words (84 times 100/7). In fact, the text contains 1500 words. Using one frequency of one word gives a rough approximation.

In this anecdotal example, we use only one word *the* to guess the real size of the text, but a better result can be obtained by using more data points in our estimation. Imagine now that we use a larger number of words, and allow each word to predict the size of the whole text. We can then average the estimations derived from each word's actual frequency to estimate more closely the size of the whole text. This is the technique that we have implemented, which we now explain in more detail, and show how accurate it is.

**Creating Language Word Count Predictors**
Over the past years, we have been accumulating texts in a wide variety of languages for work in computational linguistics in our research center. In order to create a set of predictors for these languages (as *the* in the example above is used as a predictor for the size of English texts), we proceeded as follows. For each of thirty-two different Latin-alphabet languages, we set aside 1 megabyte of training text. From this 1-megabyte training text, we extracted all alphabetic tokens[6]. We calculated the frequency of these tokens within the training text, and sorted the list of tokens by descending frequency. We retained for each language the one hundred most frequently occurring tokens with their relative frequencies. We then eliminated from each list any word appearing in more than one list (for example *que* is a high frequency word in Spanish, French, Portuguese and Catalan and was eliminated from each list). From the remaining tokens, we retained the top twenty most frequent token and their frequencies to use as predictors for that language. The predictors retained for eight languages are shown in Figure 1. For example under the heading Finnish, we find the entry *0.008 että* which means that the word *että* occurs 8 times per thousand words in the training corpus.

---
[4] http://www.excite.com
[5] The BNC is a 100 million word corpus of British English. 90 million words of English from written sources and 10 million words of transcribed spoken English. See http://info.ox.ac.uk/bnc/index.html for information.

| **English** | | **Finnish** | | **French** | | **German** | |
|---|---|---|---|---|---|---|---|
| 0.010 | that | 0.008 | että | 0.009 | pour | 0.030 | die |
| 0.010 | is | 0.008 | ei | 0.009 | dans | 0.029 | und |
| 0.008 | it | 0.006 | oli | 0.008 | une | 0.011 | zu |
| 0.007 | you | 0.005 | tai | 0.007 | qui | 0.010 | von |
| 0.007 | with | 0.005 | joka | 0.007 | par | 0.010 | mit |
| 0.007 | be | 0.005 | myös | 0.006 | sur | 0.009 | für |
| 0.006 | this | 0.004 | mutta | 0.006 | est | 0.009 | ist |
| 0.006 | are | 0.004 | ovat | 0.006 | au | 0.007 | werden |
| 0.005 | or | 0.004 | kun | 0.005 | plus | 0.007 | auf |
| 0.005 | have | 0.003 | sen | 0.004 | pas | 0.007 | eine |
| 0.005 | not | 0.003 | ole | 0.004 | sont | 0.007 | im |
| 0.005 | we | 0.003 | niin | 0.004 | aux | 0.006 | nicht |
| 0.004 | will | 0.003 | kuin | 0.004 | avec | 0.006 | ein |
| 0.004 | they | 0.003 | hän | 0.003 | cette | 0.006 | sich |
| 0.004 | all | 0.003 | voi | 0.003 | vous | 0.006 | auch |
| 0.003 | his | 0.003 | sekä | 0.003 | nous | 0.006 | oder |
| 0.003 | but | 0.003 | jos | 0.002 | ont | 0.005 | sind |
| 0.003 | your | 0.002 | vain | 0.002 | leur | 0.005 | bei |
| 0.003 | their | 0.002 | sitä | 0.002 | ses | 0.004 | als |
| 0.003 | my | 0.002 | olla | 0.002 | ces | 0.004 | wird |
| **Italian** | | **Norwegian** | | **Polish** | | **Portuguese** | |
| 0.015 | per | 0.023 | det | 0.004 | met | 0.011 | com |
| 0.014 | che | 0.018 | pê | 0.004 | jak | 0.008 | uma |
| 0.010 | della | 0.016 | til | 0.004 | przez | 0.007 | os |
| 0.007 | dei | 0.011 | har | 0.003 | lub | 0.006 | não |
| 0.006 | delle | 0.010 | jeg | 0.003 | dla | 0.003 | ao |
| 0.004 | nel | 0.009 | ikke | 0.003 | cut | 0.003 | mas |
| 0.004 | alla | 0.007 | vi | 0.002 | poland | 0.003 | muito |
| 0.004 | sono | 0.007 | om | 0.002 | dst | 0.003 | seu |
| 0.004 | ed | 0.007 | var | 0.002 | oraz | 0.002 | são |
| 0.003 | anche | 0.005 | kan | 0.002 | czy | 0.002 | vm |
| 0.003 | più | 0.005 | men | 0.002 | tylko | 0.002 | eu |
| 0.003 | gli | 0.005 | fra | 0.002 | tego | 0.002 | foi |
| 0.003 | ad | 0.005 | så | 0.002 | byc | 0.002 | você |
| 0.003 | come | 0.004 | seg | 0.002 | siê | 0.002 | ele |
| 0.003 | dal | 0.004 | dette | 0.002 | juz | 0.002 | pela |
| 0.003 | cui | 0.004 | eller | 0.002 | przy | 0.001 | quando |
| 0.002 | nella | 0.003 | vil | 0.001 | mozna | 0.001 | pode |
| 0.002 | ai | 0.003 | også | 0.001 | bardzo | 0.001 | brasil |
| 0.002 | essere | 0.003 | skal | 0.001 | tez | 0.001 | seus |

**Table 1: List of predictors and their relative frequencies for eight languages**.

---

[6] Tokenization is actually a language-dependent operation that requires language-specific resources (Grefenstette, 1999) to be done properly. We used a simple approximation to tokenization, using non-alphabetic characters to split words since this seems to be the method used on the Web browser (Altavista) that we would be exploiting later.

Once the predictors were derived for each language according to the above procedure, we created a testbed to evaluate them. For each of the eleven languages listed in the next table, Table 2, we created a 1-megabyte test text using random texts but different from those used in the training phase to derive the predictors.

**Word Count Prediction Algorithm**

Our method for using the predictors is simple. In order to estimate the number of words in a given language in a new file: (i) tokenize the new file using non-alphabetic characters as separators, (ii) sort the tokens and count their frequencies in the new file, (iii) for each predictor (we used lists of twenty predictors), divide the frequency of that token in the new file by the relative frequency of the predictor, producing that predictor's estimate of the total number of words, (iv) throw out the highest and lowest estimates (we threw out the two highest and two lowest estimates) (v) average the remaining predictions (we used the sixteen intermediate predictions) and give the average as the prediction of number of words in the new file for the language being estimated.

**Algorithm Evaluation**

To evaluate this simple method, for each 1-megabyte text in the testbed, we used the method described in the last paragraph in a first experiment to predict the known size of that text in words using the predictors for that language. The first column of Table 2 shows the variation between the real size of the testbed file and the predicted size of testbed files for the eleven languages used in this experiment. This first column shows that the predictors guess, with an error of about ten percent, the actual number of words in the testbed files for each of the eleven languages tested.

|  | Experiment 1 Only One Language | Experiment 2 All 11 TestBed Files Mixed | Experiment 3 All TestBed Files Mixed plus Noise |
|---|---|---|---|
| English | -3.8 % | +10.3 % | +13.2 % |
| Finnish | -8.1 % | -6.8 % | -6.8 % |
| French | -1.5 % | +0.9 % | +0.9 % |
| German | -7.8 % | -6.9 % | -6.9 % |
| Italian | -7.4 % | -5.7 % | -5.6 % |
| Norwegian | -7.2 % | -6.1 % | -6.1 % |
| Polish | -11.2 % | -10.1 % | -10.1 % |
| Portuguese | -6.5 % | -2.9 % | -2.9 % |
| Slovakian | -6.0 % | -4.8 % | -4.6 % |
| Slovenian | -5.7 % | -4.2 % | -3.5 % |
| Spanish | -3.2 % | +0.9 % | +1.2 % |
| *Average Error* | ±6.2 % | ±5.4 % | ±5.6 % |

**Table 2: Error in estimation of size of testbed files using predictors derived from training set. The first column shows the size estimation error of using a predictor set on a file of the same language. The second column shows the predicted size of each language in a large file containing all the testbed language files. The third column is the error rate in estimating the language size in the same file as the second column with many additional languages included.**

As a second experiment, we concatenated each of the 1-megabyte testbed files into one large 11-megabyte file composed of all eleven languages and ran each predictor over this mixed language text. This experiment is intended to show that the predictors can pick out how much of a given language is found in a mixed language text. Again, as shown in the second column of Table 2, the predictors guess within about 10% the actual number of words present in each language despite the presence of other languages.

A third experiment, whose results are also given in Table 2, adds a number of other languages to this concatenated test file as noise (we added 1-megabyte of Russian ISO5, 500 kilobytes of Thai, 700 kilobytes of Maltese). These additional languages produce little change in the estimations, except for English, which can be explained by the effective presence of English words and phrases inside these other language files.

| Size in Kb | German | Polish | Norwegian |
|---:|---:|---:|---:|
| 50 | +4.4 % | +13.8 % | +11.9 % |
| 100 | -1.0 % | +6.1 % | -6.7 % |
| 200 | -5.3 % | +12.0 % | -0.4 % |
| 300 | -5.6 % | +14.0 % | +6.9 % |
| 400 | -7.9 % | +11.5 % | +1.8 % |
| 500 | -12.2 % | +4.6 % | +2.2 % |
| 1000 | -13.4 % | +6.9 % | -4.0 % |
| 2000 | -8.1 % | -3.1 % | -4.9 % |
| 3000 | -5.0 % | -6.0 % | -5.7 % |
| 4000 | -6.3 % | -10.8 % | -5.7 % |
| 5000 | -6.3 % | -7.5 % | --- |
| 6000 | -6.7 % | -10.8 % | --- |
| 7000 | -7.4 % | -12.2 % | --- |
| 8000 | -7.1 % | -13.1 % | --- |
| **Avg.** | **±6.9 %** | **±9.5 %** | **±5.0 %** |

**Table 3: The effect on augmenting the size of a language within a 15-megabyte reference corpus of other languages. The error rates of the predictor hover around ten percent**.

In the three experiments mentioned above, each language was represented by 1-megabyte of text. In order to see how the error rate of our prediction method varies with the size of the file to be estimated, we applied the technique described above to test files of varying sizes. Figure 3 shows the variation in percent error between the real size and the guessed size for increasing amounts of German, Norwegian and Polish text mixed in with about 15 megabyte of other language text. The error rate fluctuates around the ten percent mark, with variations due to one or the other of the predictors becoming one of the outlying values that is not considered on the averaging of the estimations.

**Evaluation of Domain Specific Predictors**

The collection of files that we used for the experiments mentioned above was disparate, coming from the Web and other sources. In order to see how predictors might vary if they were created from a typed corpus of text, we generated two sets of predictors from Norwegian text. One set was derived from eclectic Norwegian texts found on the Web; the other set of predictors was derived from Norwegian legal texts. As shown in Table 4, predictors from more general texts (here from Web-based texts) are more robust than domain-dependent predictors.

|  | Error Rates | |
|---|---|---|
|  | Web Text | Non-Web Legal Text |
| Predictors from Non-Web corpus | +31.1 % | +14.3 % |
| Predictors from Web-based corpus | +4.5 % | +4.1 % |

**Table 4: Comparison between using web and non-web training corpus to create predictors. Predictors generated from disparate domains (web text) are better than domain specific predictors.**

The above controlled experiments, in which we know how big the corpora we were using were and in which we could validate our results (illustrated in Tables 2 to 4), lend confidence to the idea that, given a small number of predictor words with their relative frequencies for a reference corpus and their actual frequencies in a real corpus, we can estimate rather accurately the total number of language-specific words in the corpus. In the next section, we extend our experimentation to the WWW.

## Estimating Language Volume on the World Wide Web

In order to use the method described in the previous section over the Web, we only need to know how many times the predictors of a language occur in some indexed portion of the Web. The Web portal Altavista[7] returns two counts for each query that it is given: the number of pages that the query accesses, and the number of times each query term has been indexed by Altavista. This last number is what is needed for our estimation technique.

Here is an example of how to estimate the quantity of text available for a given language through a browser such as Altavista. Suppose that we want to estimate how many words of German have been indexed by this browser. We form a query composed of our predictor words

> **"und" "von" "für" "ist" "werden" "auf" "eine" "nicht" "sich" "auch" "oder" "sind" "wird" "aus" "einer" "durch" "wir" "daß" "wie" "zur"**

After submitting this query, Altavista responds with a page containing the line:

> **word count: daß: 7990333; durch: 8250898; einer: 9315833; wir: 9590451; wie: 9844516; wird: 11286438; sind: 11944284; zur: 12232738; oder: 13566463; aus: 13678143; auch: 15504327; werden: 16375321; sich: 17547518; nicht: 18294174; eine: 19739540; auf: 24852802; ist: 26429327; für: 33903764; von: 39927301; und: 101250806**

---

[7] http://www.altavista.com

Using the relative frequencies for these words (see Table 1), each word makes a prediction on the total number of German words available in Altavista. For example, the word *oder* has a relative frequency of 0.0056118 in our training set. And this word *oder* was found 13566463 times by Altavista (as of Feb 28, 2000). Dividing this actual frequency by the relative frequency leads to a single-word prediction of 2,417,488,684 total German words accessible through Altavista. Applying the same technique to the word *und* yields a prediction of 3,500,617,348 German words accessible through Altavista. Throwing out the highest and lowest estimates and averaging the rest yields an estimation of about 3 billion words of German accessible through Altavista in February 2000.

| predictor | Relative Frequency | Single Word Prediction |
|---|---|---|
| … | … | … |
| oder | 0.00561180 | 2417488684 |
| sind | 0.00477555 | 2501132644 |
| auch | 0.00581108 | 2668062907 |
| wird | 0.00400690 | 2816750605 |
| nicht | 0.00646585 | 2829353294 |
| eine | 0.00691066 | 2856389983 |
| sich | 0.00604594 | 2902363900 |
| ist | 0.00886430 | 2981546991 |
| auf | 0.00744444 | 3338438082 |
| und | 0.02892370 | 3500617348 |
| … | … | … |
| **Average Prediction** | | 3,068,760,356 |

**Table 5: Estimation of number of German words available through Altavista using German predictors.**

In February 2000, we constructed Altavista queries corresponding to the predictors, such as those given in Figure 1, for the 32 Latin character languages in Table 6. These results are roughly consistent with Jack Xu (2000) page counts in that his team at Excite found that 71% of the pages (including non-Latin language pages) that they trawl are in English, and 5.1% were in German, giving a ratio of 14 to 1 of English to German. In our word counts, we estimate a similar relation of 15 to 1 based on word count predictions. The order of other Latin-alphabet languages also follows that of the Excite study: French, Spanish, Italian, (Portuguese), Swedish, …

|              | *Word count estimate* |
|--------------|----------------------:|
| **Welsh**    | 7,590,000             |
| **Albanian** | 9,203,000             |
| **Breton**   | 9,975,000             |
| **Lithuanian** | 20,927,000          |
| **Latvian**  | 21,925,000            |
| **Esperanto**| 26,795,000            |
| **Basque**   | 28,296,000            |
| **Latin**    | 38,256,000            |
| **Estonian** | 43,257,000            |
| **Irish**    | 49,778,000            |
| **Icelandic**| 53,167,000            |
| **Roumanian**| 63,846,000            |
| **Croation** | 72,122,000            |
| **Slovenian**| 74,998,000            |
| **Turkish**  | 100,548,000           |
| **Malay**    | 113,236,000           |
| **Catalan**  | 126,324,000           |
| **Slovakian**| 140,909,000           |
| **Finnish**  | 192,105,000           |
| **Danish**   | 206,167,000           |
| **Polish**   | 235,726,000           |
| **Hungarian**| 268,944,000           |
| **Czech**    | 269,310,000           |
| **Norwegian**| 453,391,000           |
| **Dutch**    | 622,063,000           |
| **Swedish**  | 644,740,000           |
| **Portuguese**| 924,965,000          |
| **Italian**  | 1,240,205,000         |
| **Spanish**  | 1,595,489,000         |
| **French**   | 2,208,418,000         |
| **German**   | 3,068,760,000         |
| **English**  | 47,264,700,000        |

**Table 6: Estimation of number words available for a number of Latin character languages through the Altavista web portal.**

**Historical Estimations**

We captured data for predicting word counts at three points in time: end October 1996, mid August 1999 and end February 2000. Table 7 shows the evolution of eight European languages through the Altavista portal, as estimated by a set of word count predictors we had first used in 1996[8]. From the ratio progressions from 1996 to 1999, we see that although the quantity of English that Altavista is indexing is growing quickly, the volumes of other languages than English are growing at faster pace than that of English.

---

[8] The predictors used here in Table 7, the same ones for which we gathered data in 1996, are slightly different from thos used to calculate the estimate given in Table 6, which explains the discrepancies in the estimates between the two tables.

|  | Oct 1996 | Ratio to English | Aug 1999 | Ratio to English | Feb 2000 | Ratio to English |
|---|---|---|---|---|---|---|
| English | 6,082,090,000 | 1.000 | 28,222,100,000 | 1.000 | 48,064,100,000 | 1.000 |
| German | 228,938,428 | 0.038 | 1,994,229,409 | 0.071 | 3,333,127,671 | 0.069 |
| French | 223,316,023 | 0.037 | 1,529,795,169 | 0.054 | 2,732,221,327 | 0.057 |
| Spanish | 104,319,158 | 0.017 | 1,125,646,460 | 0.040 | 1,894,966,981 | 0.039 |
| Italian | 123,555,682 | 0.020 | 817,270,444 | 0.029 | 1,338,351,674 | 0.028 |
| Portuguese | 106,167,245 | 0.017 | 589,391,943 | 0.021 | 1,161,898,076 | 0.024 |
| Norwegian | 106,497,066 | 0.018 | 669,331,120 | 0.024 | 947,486,593 | 0.020 |
| Finnish | 20,647,404 | 0.003 | 107,260,274 | 0.004 | 166,599,467 | 0.003 |

**Table 7: Estimation of language size in 1996, late 1999 and early 2000. English has grown 800% over this period, but German has grown 1500%, and Spanish has grown 1800% in the same period.**

## Conclusions

In this paper we described a method for estimating the word count of a specific language within a corpus, given only the frequencies of a few words in the corpus. We tested this method by applying it to known corpora, both mono and multilingual, measuring its reliability. We then applied it to frequency data obtained from a popular web portal, Altavista. Results of this procedure give an indication of language volume on the part of the Web that is visited by Altavista. Applying the same technique to data gathered over three years gives an idea of the evolution of language volume for some European languages from 1996 to 2000. We can be reasonably sure that these estimates of the number of words accessible from this browser for these languages are true to within about 10%. But according to Steve Lawrence and Lee Giles, major search engines including Altavista, seem to cover only about 16% of the indexable web, and are more likely to index commercial and US sites. Our results must be interpreted in light of this. It is not sure just what 16% of the Web is represented by Altavista's web trawling: whether it is evenly distributed sampling or biased to North America. But the relative frequency and relative growth of languages within this sample can be estimated by the technique described here, which thus provides a tool for measuring this linguistic characteristic of the Web.